\documentclass[runningheads]{llncs}
\usepackage{graphicx}
\usepackage{amsmath}
\usepackage[T1]{fontenc}
\usepackage{lmodern}

\usepackage{amssymb}
\usepackage{subcaption}
\usepackage{enumitem}
\usepackage{comment}
\usepackage{url}
\usepackage{xcolor}
\usepackage{capt-of}
\usepackage{etoolbox}
\usepackage[utf8]{inputenc} % allow utf-8 input
\usepackage{hyperref}       % hyperlinks
\usepackage{url}            % simple URL typesetting
\usepackage{booktabs}       % professional-quality tables
\usepackage{amsfonts}       % blackboard math symbols
\usepackage{nicefrac}       % compact symbols for 1/2, etc.
\usepackage{microtype}      % microtypography
\usepackage{xcolor}         % colors
\usepackage{balance}

\setlist{leftmargin=*}

\usepackage{algorithm}
\usepackage{algpseudocode}
\newcommand{\bfx}{{\bf x}}

\usepackage[accsupp]{axessibility}  % Improves PDF readability for those with disabilities.
\usepackage{booktabs}

\begin{document}
% \linenumbers
\pagestyle{headings}
\mainmatter
\def\ECCVSubNumber{7847}  % Insert your submission number here

\title{MINER:  Multiscale Implicit Neural Representation} % Replace with your title

%\begin{comment}
\titlerunning{ECCV-22 submission ID \ECCVSubNumber} 
\authorrunning{ECCV-22 submission ID \ECCVSubNumber} 
\author{Anonymous ECCV submission}
\institute{Paper ID \ECCVSubNumber}
%\end{comment}
%******************

\titlerunning{MINER}
\author{Vishwanath Saragadam, Jasper Tan, Guha Balakrishnan,\\ Richard G.~Baraniuk, Ashok Veeraraghavan}
\authorrunning{MINER, Saragadam et al.}
\institute{Rice University, Houston TX 77005, USA \\
\email{vishwanath.saragadam@rice.edu}}

%******************
\maketitle

\begin{abstract}
We introduce a new neural signal model designed for efficient high-resolution representation of large-scale signals. 
The key innovation in our {\em multiscale implicit neural representation} (MINER) is an internal representation via a Laplacian pyramid, which provides a sparse multiscale decomposition of the signal that captures orthogonal parts of the signal across scales. We leverage the advantages of the Laplacian pyramid by representing small disjoint patches of the pyramid at each scale with a small MLP. This enables the capacity of the network to adaptively increase from coarse to fine scales, and only represent parts of the signal with strong signal energy. The parameters of each MLP are optimized from coarse-to-fine scale which results in faster approximations at coarser scales, thereby ultimately an extremely fast training process. We apply MINER to a range of large-scale signal representation tasks, including gigapixel images and very large point clouds, and demonstrate that it requires fewer than 25\% of the parameters, 33\% of the memory footprint, and 10\% of the computation time of competing techniques such as ACORN to reach the same representation accuracy. A fast implementation of MINER for images and 3D volumes is accessible from \url{https://vishwa91.github.io/miner}.

\end{abstract}

\section{Introduction}\label{section:intro}
Neural implicit representations have emerged as a promising paradigm for signal representation and interpolation with pervasive applications in 3D view synthesis~\cite{mildenhall2020nerf,peng2020convolutional,kuznetsov2021neumip,srinivasan2021nerv}, images~\cite{chen2021learning}, video~\cite{chen2021nerv}, and linear inverse problems~\cite{sun2021coil,chen2021learning}.
At the core of such neural representations is one or several multi layer perceptrons (MLPs) that produce a continuous mapping from signal coordinates to the values of the signal at those coordinates.

The success of neural implicit representations relies on the ability to fit models accurately (high representation accuracy), rapidly (short training time), and in a concise manner (small number of parameters).
However, most state-of-the-art implicit representations require training a single large MLP (parameters in millions) that suffers from high computational cost, requiring large memory footprints and long training times. 
While there have been several modifications to the network architecture~\cite{sitzmann2020implicit,reiser2021kilonerf,martel2021acorn} and inference~\cite{yu2021plenoctrees}, neural implicit representations are not yet practical for handling extremely high dimensional signals such as gigapixel images or 3D point clouds with several billion data points.

\begin{figure*}[!tt]
	\centering
	\includegraphics[width=\textwidth]{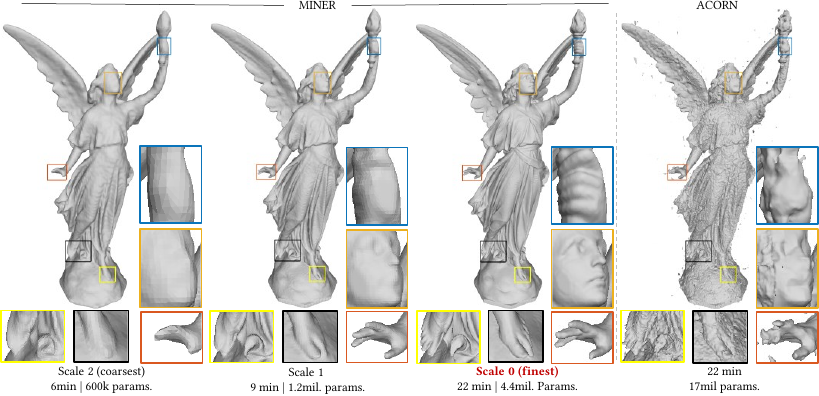}
	\caption{\textbf{Multiscale Implicit Representations.} We present a novel implicit representation framework called MINER that is well suited for very large visual signals such as images, videos, and 3D volumes. We leverage the self-similarity of visual signals across scales to iteratively represent models from coarse to fine scales, resulting in a dramatic decrease in inference and training time, while requiring fewer parameters and less memory than state-of-the-art representations. This figure demonstrates fitting of the Lucy 3D mesh over three scales with scale 2 being the coarsest and 0 being the finest. MINER achieves high quality results across all scales with high IoU value and achieves an IoU of 0.999 at the finest scale in less than 30 minutes. In comparison, the state-of-the-art approach (ACORN) results in an IoU of 0.97 in that time, while requiring far more parameters.}
	\label{fig:teaser}
\end{figure*}

\begin{figure*}[!tt]
	\centering
	\includegraphics[width=\textwidth]{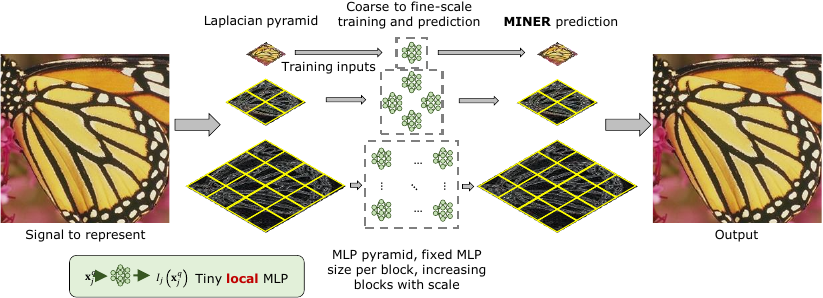}
	\caption{\textbf{MINER trains and predicts Laplacian pyramids.} Visual signals are similar across scales and are compactly represented by Laplacian pyramids. MINER follows a similar scheme where each scale of the Laplacian pyramid is represented by multiple, local MLPs with a small number of parameters. The number of such MLPs increase by a factor of 2 from coarse to fine scale, thereby representing a fixed spatial size at each scale. This multi-scale representation naturally lends itself to a sequential, coarse-to-fine scale training process that is fast and memory efficient.}
	\label{fig:representation}
\end{figure*}

We introduce a multiscale implicit neural representation (MINER) that is well-suited for representing very high dimensional signals in a concise manner.
Our key observation is that Laplacian pyramids of visual signals offer a sparse and multiscale decomposition that naturally separates a signal's frequency content across spatial scales.
We leverage the multiscale decomposition by representing each spatial scale of the Laplacian pyramid with  different MLPs. 
Instead of using a single MLP at each scale, we represent a small disjoint image/volume patch of fixed size with a small MLP, resulting in both a multiscale and multipatch decomposition.
Such a multipatch decomposition is well-suited for sparse signals as most patches will have near-zero intensity, thereby not requiring an explicit MLP for that patch.
MINER enables a fast and flexible multi-resolution analysis, as representing the signal at lower resolution requires training $2\times$ fewer MLPs along each spatial dimension (due to fewer patches), 
An example on fitting a 3D volume across three spatial scales on one billion points is shown in Fig.~\ref{fig:teaser}.
MINER provides a visually pleasing result even at the coarsest spatial scale in six minutes with as few as 600k parameters. The finest scale converges in 22 minutes. In contrast, for the same amount of training time, state-of-the-art approaches such as ACORN result in many artifacts while also requiring $4\times$ more parameters.
An overview of the MINER signal model is shown in Fig.~\ref{fig:representation}.

The multiscale, multi-MLP architecture lends itself to a fast and memory efficient training procedure.
At each spatial scale, the parameters of the MLPs are trained for the corresponding Laplacian pyramid decomposition.
We then \emph{sequentially} train MLPs from coarsest scale to the finest scale.
The near-orthogonality of the Laplacian transform across scales ensures that new information is added at every scale, thereby resulting in an iterative refinement framework.
We leverage the sparsity of the Laplacian transform by comparing the upsampled signal from the fine block and the target signal at that fine block -- if the error in representation (or variance of signal) is smaller than a threshold, we prune out the blocks before training starts.
This leads to fewer blocks to train at finer resolutions.

MINER is $10\times$ or more faster compared to state-of-the-art implicit representations in terms of training process for a comparable number of parameters and target accuracy.
MINER can represent gigapixel images with greater than 38dB accuracy in less than three hours, compared to more than a day with techniques such as ACORN~\cite{martel2021acorn}.
For 3D point clouds, MINER achieves an intersection over union (IoU) of $0.999$ or higher in less than three minutes, resulting in two orders of magnitude speed up over ACORN.
Due to the multiscale representation, MINER can be used for streaming reconstruction of images, as with JPEG2000~\cite{shapiro2009embedded}, or efficiently sampling for rendering purposes with octrees~\cite{yu2021plenoctrees} -- making neural representations ready for extremely large scale visual signals.

\section{Prior Work}\label{section:prior}
MINER draws inspiration from classical multiscale techniques and more recent neural representations. We outline some of the salient works here to set context.
\paragraph{Implicit neural representations.} Implicit neural representations learn a continuous mapping from local coordinates to the signal value such as intensity for images and videos, and occupancy value for 3D volumes.
%For images, implicit representations learns the intensity at each pixel location, $\mathcal{N}:(x, y) \rightarrow \text{im}(x, y)$, where $\mathcal{N}$ is a multi-layer perceptron (MLP). For videos, this would be, $\mathcal{N}: (x, y, t) \rightarrow V(x, y, t)$.
%
%The goal is then to learn the parameters of the network $\mathcal{N}$ that maximizes the representation accuracy.
%
The learned models are then used for a myriad of tasks including image representations~\cite{chen2021learning}, multi-view rendering~\cite{mildenhall2020nerf}, and linear inverse problems solving~\cite{sun2021coil}.
Recent advances in the choice of coordinate representation~\cite{tancik2020fourfeat} and non-linearity~\cite{sitzmann2020implicit} have resulted in training processes that have high fitting accuracy.
Salient works related to implicit representations include the NeRF representations~\cite{mildenhall2020nerf} and its many derivatives that seek to learn the 3D geometry from a set of multi-view images.
Despite the interest and success of these implicit representations, current approaches often require disproportionately large number of parameters compared to the signal dimension.
This culminates in a large memory footprint and training times, precluding representation of very high-dimensional signals.
%
%The major drawback of existing implicit representation is the training time, which often extends to multiple days.

\paragraph{Architectural changes for faster learning.}
Several interesting modifications have been proposed to increase training or inference speed. KiloNeRF~\cite{reiser2021kilonerf} and deep local shapes~\cite{chabra2020deep} replaced the large MLP with multiple small MLPS that fit only a small, disjoint part of the 3D space. 
Such approaches dramatically speed up the inference time (often by $60\times$) and in some cases enable better generalization~\cite{mehta2021modulated}, but they have little to no effect on the training process itself.
ACORN~\cite{martel2021acorn} utilized an adaptive coordinate decomposition to efficiently fit various signals.
%
%The underlying idea of ACORN is to decompose the signal into small disjoint regions with approximately equal information.
%
%The decomposition ensured that regions with more texture were enclosed in small blocks and regions with little to no texture were enclosed in large blocks. 
By utilizing a combination of integer programming and interpolation, ACORN reduced training time for fitting of images and 3D point clouds by one to two orders of magnitude compared to techniques like SIREN~\cite{sitzmann2020implicit} and the convolutional occupancy network~\cite{peng2020convolutional}.
However, ACORN does not leverage the cross-scale similarity of visual signals, and this often leads to long convergence times for very large signals. 
Moreover, the adaptive optimized blocks requires integer programming and several hundreds of thousands of gradient  steps which can be prohibitively expensive.

\paragraph{Multi-scale representations.} Visual signals are similar across scales, and this has been exploited for a wide variety of applications. 
In computer vision and image processing, the wavelet transform and Laplacian pyramids are often used to efficiently perform tasks such as image registration~\cite{thevenaz2000optimization}, optical flow computation~\cite{weber1995robust}, and feature extraction~\cite{lowe2004distinctive}.
Multi-scale representations such as octrees~\cite{meagher1982geometric,chien1986volume} and mip-mapping are used to speed up the rendering pipeline. %Similarly, MIP-mapping utilizes multi-scale texture representation for faster rendering, as well as removing aliasing artifacts.
This has also inspired neural mipmapping techniques~\cite{kuznetsov2021neumip} that utilize neural networks to represent texture at each scale.
Along the same lines, spatially adaptive progressive encoding~\cite{hertz2021sape} enables a coarse-to-fine training approach that gradually learns higher spatial frequencies.
Multi-scale representations are also utilized for several linear inverse problems such as multi-scale dictionary learning for denoising~\cite{sulam2014image}, compressive sensing~\cite{park2009multiscale}, and sparse approximation~\cite{mairal2007multiscale}.
Some recent works have focused on a level-of-detail approach to neural representations~\cite{takikawa2021neural} (NGLOD) where the multiple scales are \emph{jointly} learned. 
The implicit displacement fields (IDF) approach~\cite{yifan2021geometry} similarly learns a smooth approximation of the surface, along with a high frequency displacement at each spatial point to represent the shape.
While efficient in rendering, NGLOD and IDF have no advantage in the training phase, as it relies on training all levels of detail at the same time.
MINER also results in an LOD representation, but the underlying approach is significantly different.
MINER relies on a block-wise representation at each scale with sequential training from coarse to fine scales, which enables more compact representation with faster training times.
%endows it with several advantages including cross-scale weight sharing, early termination, and fast and efficient training.
%
%Table~\ref{tab:comparison} compares various methods in terms of MLP architecture, how signal is split, and training and inference times.

%\begin{table}[!tt]
%	\centering
%	\caption{\textbf{Comparison of various methods.} \vishwa{supplementary.} Existing MLP architectures fit signal at a single scale with a single large MLPs or multiple small MLPs. MINER fits signals at multiple scales with multiple small MLPs which results in extremely fast training and inference times.}
%	\includegraphics[width=\columnwidth]{figures/comparison_tables.pdf}
%	\label{tab:comparison}
%\end{table}

\section{MINER}\label{section:miner}
%
%MINER draws inspiration from KiloNeRF~\cite{reiser2021kilonerf} in terms of representing small parts of the signal with a small MLP instead of a single large MLP.
%
%However, MINER goes beyond a KiloNeRF-like representation by fitting the signal at multiple scales.
%
%We first describe the representation for 2D images and then show how it can be extended to 3D volumes.

MINER leverages a block decomposition of signal along with a Laplacian pyramid-like representation. We now describe the MINER signal model and the training process.

\subsection{Signal model}
Let $\bfx$ be the coordinate and $\text{I}(\bfx)$ be the target. We will assume that the coordinates lie in $[-1, 1]$. 
Let $\mathcal{D}_{j}$ be the domain specific operator that downsamples the signal by $j$ times, and $\mathcal{U}_j$ be the domain-specific operator that upsamples the signal by $j$ times.
%
%We define the downsampling and upsampling operators as follows:
%\begin{align}
%	\mathcal{D}_{j}~&:~\mathbb{R}^{H\times W} \mapsto \mathbb{R}^{H/2^{j} \times W/2^j}&&\text{(Downsampler)}\\
%	\mathcal{U}_{j}~&:~\mathbb{R}^{H/2^j \times W/2^j} \mapsto \mathbb{R}^{H \times W}&&\text{(Upsampler)}.
%\end{align}
%
%As an example, an \emph{area} downsampling operation for images would be,
%\begin{align}
%    \mathcal{D}_j(I) = I_j(\bfx_j) = %\int_{0}^{2^j/H}\int_{0}^{2^j/W} %\text{I}(2^j\bfx_j + (dx, dy)) dx dy.
%\end{align}
%
We will leverage $J$ implicit representations, $I_j (\bfx) \approx N_j$ for \mbox{$j \in [0,J-1]$}, where $N_j$ is the MLP at the $j^\text{th}$ level of a Laplacian pyramid, a multiscale representation which separates the input signal into scales capturing unique spatial frequency bands. Two desirable properties of such a bandpass pyramid is that signals across scales are approximately orthogonal to one another~\cite{do2003framing} and are sparse. We found in our experiments that these properties dramatically reduce the training and inference times compared to a lowpass pyramid such as the Gaussian pyramid (see Fig.~\ref{fig:signal_vs_residue}). 

%We specifically use a representation based on a Laplacian pyramid, due to its known self-similarity property across scales [CITE?].
%\begin{align}
%	I_j (\bfx) \approx N_j %(\bfx)\label{eq:signal_rep},
%\end{align}

%
%Although such a representation is multiscale by definition, the self-similarity across scales is not exploited.
%\begin{figure}[!tt]
%	\centering
%	\includegraphics[width=\columnwidth]{figures/why_kilonerf.pdf}
%	\caption{\textbf{Multiple MLPs allow a more compact representation.} The figure above visualizes how MINER approximates a signal like a Laplacian pyramid. Several blocks (marked in red) have a very low residue strength, implying that they can be represented simply as a zero signal. We leverage this observation by having one MLP per \emph{each active block}, thereby requiring fewer parameters, and ultimately simpler inference.}
%	\label{fig:why_kilonerf}
%\end{figure}
%For a more concise representation that leverages cross-scale similarity, we follow a Laplacian pyramid-like representation,
Letting $R_j$ denote the MLP modeling the residual signal at scale $j$, our Laplacian pyramid representation may be written as:
\begin{align}
	I_{J-1}(\bfx) &=\mathcal{D}_{J-1}(I)(\bfx)\approx N_{J-1} (\bfx)\label{eq:coarsest}\\
	I_{J-2}(\bfx) &=\mathcal{D}_{J-2}(I)(\bfx)\approx R_{J-2}(\bfx) + \mathcal{U}_2(N_{J-1}(\bfx/2))\label{eq:finer}\\
	&\vdots\nonumber\\
	I(\bfx) &\approx R_0(\bfx) + \mathcal{U}_2(N_1(\bfx/2))\\
	%		&\approx R_0(\bfx) + %\mathcal{U}_2(R_1(\bfx/2)) + %\mathcal{U}_4(N_2(\bfx/4))\\			
	%I(\bfx) 
	&\approx N_0(\bfx) + \mathcal{U}_2(N_1(\bfx/2)) + \cdots +
	   \mathcal{U}_{2^{J-1}}(N_{J-1}(\bfx/2^{J-1}))\label{eq:finest},
\end{align}

where eq.~\eqref{eq:coarsest} is the coarsest representation of the signal. At finer scales (as in eq.~\eqref{eq:finer}), we write the signal to be approximated as a sum of the upsampled version of the previous scale and a residual term. This results in a recursive multi-resolution representation that naturally shares information across scales. %as opposed to the purely signal-based representation in eq. \eqref{eq:signal_rep}.

We make two observations about MINER:
\begin{itemize}
	\item Signals at coarser resolutions are low-dimensional, and therefore require smaller MLPs. These MLPs are faster for inference, which is beneficial for tasks such as mipmapping and LOD-based rendering.
	
	\item The parameters of the MLPs up to scale $j$ only rely on the signal at scales $q = j, j+1, \cdots, J-1$. This implies the MLPs can be trained \emph{sequentially} from coarsest to finest scale. We will see next that this offers a dramatic reduction in training time without sacrificing quality.
\end{itemize}

\subsubsection{Using multiple MLPs per scale.} Equation \eqref{eq:finest} implies that obtaining a value at a spatial point $\bfx$ requires evaluating a total of $J$ MLPs across scales. Such joint evaluation has no computational benefit compared to a single scale approach like SIREN~\cite{sitzmann2020implicit} with comparable number of parameters. Further, the residual signals at finer scales are often low-amplitude, a consequence of visual signals being composed of several smooth areas. To leverage this fact and make inference faster, we split the signal into equal sized blocks at each scale. We create an MLP for each block that requires significantly fewer parameters than a single full MLP at that scale. Moreover, blocks with small residual energy can be represented as a zero signal, and do not even need to be represented with an MLP.

We now combine the Laplacian representation with the multi-MLP approach stated above. Let $\widetilde{\bfx}$ be a local coordinate at the finest scale in a block with coordinate $(m, n)$, where $m \in {1, 2, \cdots M}$ is the number of vertical blocks, and $n \in {1, 2, \cdots, N}$ is the number of horizontal blocks. To evaluate the signal at $\bfx$,
\begin{align}
	I(\bfx) &= I\left(\widetilde{\bfx} + \left[\frac{mH}{M}, \frac{nW}{N}\right]\right)
			= \mathcal{R}^{(m, n)}_0\left(\widetilde{\bfx} + \left[\frac{mH}{M}, \frac{nW}{N}\right]\right) + \cdots \nonumber\\
			&+ \cdots \mathcal{U}_2\left(\mathcal{N}^{(\lfloor m/2\rfloor, \lfloor n/2\rfloor)}_1\left(\widetilde{\bfx} + \left[\left\lfloor\frac{mH}{2M}\right\rfloor, \left\lfloor\frac{nW}{2N}\right\rfloor\right]\right)\right),
\end{align}
where $\lfloor \cdot \rfloor$ is the floor operator, and $\mathcal{N}^{(m, n)}_j$ is the MLP for block at $(m, n)$ and at scale $j$. With this formulation, we require evaluation of at most $J$ \emph{small} MLPs instead of large MLPs, thereby dramatically reducing inference time. %Moreover, for parts of the signal with low residual energy, we simply set the value to 0 which is a constant time operation. 
%We call this representation MINER for Multiscale Implicit NEural Representation.
%
%Armed with the MINER signal representation, we will now propose a fast and memory efficient training procedure.

\subsection{Training MINER}
MINER requires estimation of parameters at each scale and each block. We now present an efficient \emph{sequential} training procedure that starts at the coarsest scale and trains up to the finest scale.

\paragraph{Training at coarsest scale.} The training process starts by fitting $\text{I}_{J-1}(\bfx)$, the image at the coarsest scale. 
We estimate the parameters of each of the MLPs $\mathcal{N}^{(m, n)}_{J-1}$ by solving the objective function,
\begin{align}
    \min_{\mathcal{N}_{J-1}^{(m, n)}} \left\| I_{J-1}\left(\widetilde{\bfx} + \left(\frac{mH}{2^{J-1}M}, \frac{nW}{2^{J-1}N}\right)\right) - \mathcal{N}_{J-1}^{(m, n)} (\widetilde{\bfx})\right\|^2.
\end{align}
Let $\widehat{I}_{J-1}(\bfx_{J-1})$ be the estimate of the image at this stage.
%
%We will detail the size of each MLP, and the optimization parameters in the experiments section.

\paragraph{Pruning at convergence.} As the training proceeds, some MLPs, particularly for blocks with limited variations, will converge to a target mean squared error (MSE) earlier than the more complex blocks. We remove those MLPs that have converged from the optimization process and continue with the remaining blocks.

\paragraph{Training at finer scales.}
%
%However, we exploit information from the previous scale for a more efficient training procedure.
%
%Inspired by Laplacian pyramids~\cite{burt1987laplacian}, we fit to the difference of the upsampled image from previous %scales and the signal at current scale.
%
As with the coarsest scale, we continue to fit small MLPs to blocks at each finer scale. For scale $J-2$, the target signal is given by
\begin{align}
    R_{J-2}(\bfx) = \text{I}_{J-2}(\bfx) -  \mathcal{U}_2(\widehat{\text{I}}_{J-1})(\bfx/2).
\end{align}
We leverage the fact that blocks within each scale occupy disjoint regions and can optimize each MLP independently of one another:

\begin{align}
	\min_{\mathcal{R}_{J-2}^{(m, n)}} \left\| R_{J-2}\left(\widetilde{\bfx} + \left(\frac{mH}{2^{J-2}M}, \frac{nW}{2^{J-2}N}\right)\right) - \mathcal{R}_{J-2}^{(m, n)} (\widetilde{\bfx})\right\|^2.
\end{align}

 %Sequential training of MLPs at each scale implies that we can rapidly get a good fit at lower resolution. 

\paragraph{Pruning before optimization.} Due to the sparseness of gradients of visual signals, we expect a large number of spatial regions to have little to no signal. Nominally, the number of blocks and MLPs double along each dimension at finer scales. However, some blocks may already be adequately represented by the corresponding MLP at the coarser scale. In such a case, we do not assign an MLP to that block and set the estimate of the residue to all zeros. Depending on the frequency content in the image, this decision dramatically reduces the number of total MLP parameters, and thereby the overall training and inference times.
In cases where \textit{a priori} information about residual energy is not available (such as view synthesis from images), we can rely on each block's variance. A block with low variance is likely to have converged at the coarser scale and hence can be pruned from training.
\begin{figure}[!tt]
    \centering
    \begin{subfigure}[t]{0.46\columnwidth}
    	\centering
    	\includegraphics[width=\columnwidth]{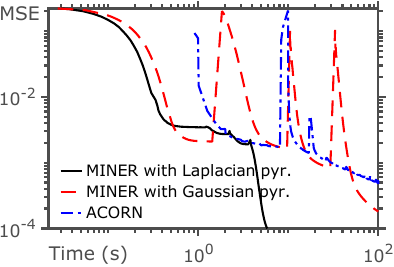}
    	\caption{Laplacian vs. Gaussian pyramid}
    	\label{fig:signal_vs_residue}
    \end{subfigure}
	\begin{subfigure}[t]{0.48\columnwidth}
		\centering
		\includegraphics[width=\columnwidth]{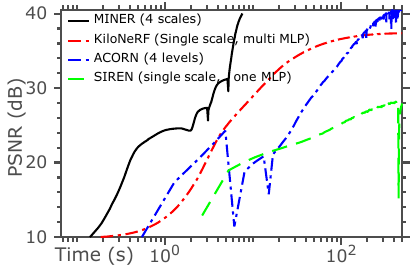}
		\caption{MINER vs. state-of-art}
		\label{fig:imfit_timing}
	\end{subfigure}    
    \caption{\textbf{Laplacian pyramid enables faster convergence.} The plot in (a) shows training error across time for a $2048\times2048$ image of Pluto. MINER when combined with a Laplacian pyramid offers a significantly faster convergence as the MLPs at finer scale capture orthogonal information compared to the coarser ones. This also results in very small jumps in training accuracy that is strongly prevalent when MINER is trained with a Gaussian pyramid, or ACORN.  The plot in (b) PSNR as a function of time for various approaches for a one megapixel image. MINER achieves higher accuracy at all times, and converges significantly faster than competing approaches. Moreover, the drop in accuracy when changing from coarse to fine scale is less severe for MINER compared to when ACORN re-estimates coordinate decomposition.}
%    \label{fig:signal_vs_residue}
\end{figure}

\section{Experimental Results}\label{section:experiments}
\paragraph{Baselines.} For fitting to images and 3D volumes, we compared MINER against SIREN~\cite{sitzmann2020implicit}, KiloNeRF~\cite{reiser2021kilonerf}, and ACORN~\cite{martel2021acorn}.
We also compared MINER Against convolutional occuppancy networks~\cite{peng2020convolutional} for 3D volumes.
%
%\begin{enumerate}
%    \item SIREN~\cite{sitzmann2020implicit} fits a single large MLP at a single scale and utilizes a sinusoidal activation function for accelerated training. We varied the number of hidden units for each experiment to ensure that the number of parameters matched that of MINER.
%    \item KiloNeRF~\cite{reiser2021kilonerf} fits multiple small MLPs at a single scale instead of a single large MLP. The number of hidden units for each MLP was chosen to be the same as that for MINER.
%    \item ACORN~\cite{martel2021acorn} fits a single large MLP at a single scale with adaptive coordinate decomposition. 
%    \item Convolutional occupancy network~\cite{peng2020convolutional} utilizes convolutions to capture local correlations. We used this only for 3D volume comparisons.
%\end{enumerate}
%
We used code from the respective authors and optimized the training parameters for a fair comparison.

\paragraph{Training details.} We implemented MINER with the PyTorch~\cite{NEURIPS2019_9015} framework.
Multiple MLPs were trained efficiently using the block matrix multiplication function (\verb*|torch.bmm|) and hence, we required no complex coding outside of stock PyTorch implementations.
All our models were trained on a system unit equipped with Intel Xeon 8260 running at 2.4Ghz, 128GB RAM, and NVIDIA GeForce RTX 2080 Ti with 12GB memory. 
For all experiments, we excluded any time taken by logging activities such as saving models, images, meshes, and computing intermittent metrics such as PSNR and IoU.
%

%\begin{figure*}[!tt]
    %\centering
   % \includegraphics[width=\textwidth]{figures/pluto.pdf}
  %  \caption{\textbf{Image fit over time.} \vishwa{supplementary.} The figure compares fitting of the 16 megapixel pluto image at various times during the training process. A distinct advantage of MINER is that the signal is similar to the final output (albeit downsampled) from the starting itself which enables an easy visual debug of the fitting process.}
 %   \label{fig:imcompare}
%\end{figure*}

\paragraph{Fitting images.}
We split up RGB images into $32\times32\times3$ patches at all spatial scales. For each patch and at each scale, we trained a single MLP with two hidden layers and sinusoidal activation function~\cite{sitzmann2020implicit}. We fixed the number of features to be $20$ for each layer. We did not add any further positional encoding.
We used the ADAM~\cite{kingmaB14} optimizer with a learning rate of $5\times10^{-4}$ and an exponential decay with $\gamma = 0.999$.
At each scale, we trained either for 500 epochs, or until the change in loss function was greater than $2\times10^{-7}$.
%
%
%The number of scales were variable -- for images smaller than 1 megapixel, we chose 3 scales.
%
%For larger images up to 64 megapixels, we chose 5 scales.
%
We used an $\ell_2$ loss function at all scales with no additional prior.
We pruned a block from the training pipeline if the block MSE was smaller than $10^{-4}$. 
Similarly, a block was not added at the starting of the training process if the block MSE was smaller than $10^{-4}$
The effect of block-stopping threshold is analyzed  in the supplementary material.

\begin{figure*}[!tt]
    \centering
    \includegraphics[width=\textwidth]{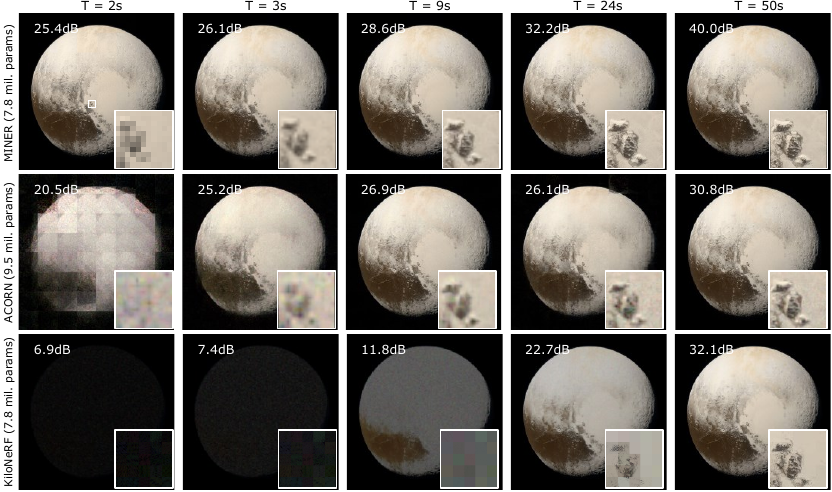}
    \caption{\textbf{Image fit over time.} The figure compares fitting of the 16 megapixel pluto image at various times during the training process. A distinct advantage of MINER is that the signal is similar to the final output (albeit downsampled) from the starting itself which enables an easy visual debug of the fitting process.}
    \label{fig:imcompare}
\end{figure*}

Figure~\ref{fig:signal_vs_residue} shows training error for a 4 megapixel (MP) image of Pluto across epochs for MINER by representing a Laplacian pyramid, a Gaussian pyramid, and ACORN.
MINER converges rapidly to an error of $10^{-4}$ compared to other approaches.
Moreover, the periodic and abrupt increase in error are more prevalent in Gaussian representation, and ACORN, which further hamper their performance, but not with Laplacian pyramid due to near-orthogonality of signal across scales.
Figure~\ref{fig:imfit_timing} shows training error for a 1 MP image for various approaches with a fixed number of parameters (900k). MINER with four scales is nearly two orders of magnitude faster than all approaches.
Figure~\ref{fig:imcompare} shows the fitting result for a 64 megapixel Pluto image across training iterations. 
The times correspond to the instances when MINER converged at a given scale. 
MINER maintains high quality reconstruction at all instances due to the multiscale training scheme and rapidly converges to a PSNR of 40dB within 50 seconds.
In contrast, ACORN achieves qualitatively good results after 10s and achieves a PSNR of $30.8$ dB after 50s, and KiloNeRF achieves a qualitatively good result only after 50s.
SIREN Results are not shown in the plot as the first epoch was completed after 4 minutes.
Results with analysis on effect of parameters such as number of scales and patch size is included in supplementary.

\begin{figure}[!tt]
    \centering
    \includegraphics[width=0.49\columnwidth]{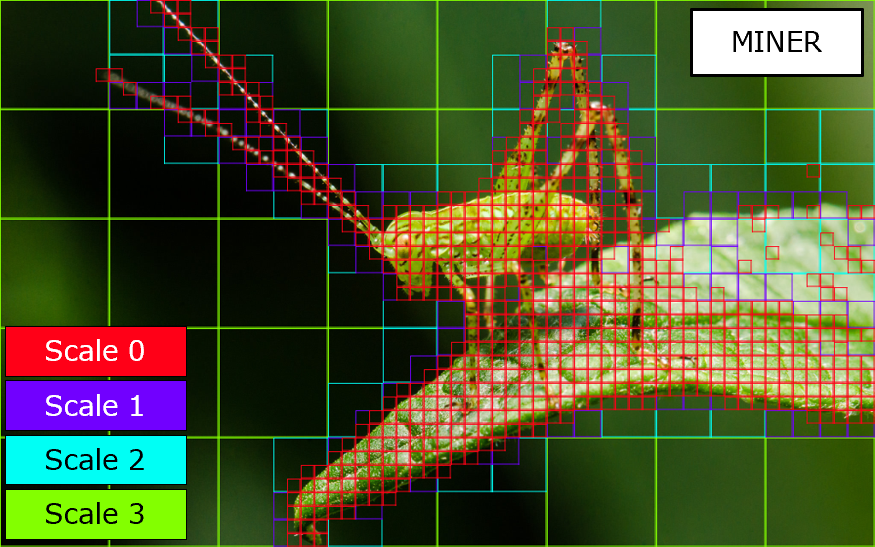}
    \includegraphics[width=0.49\columnwidth]{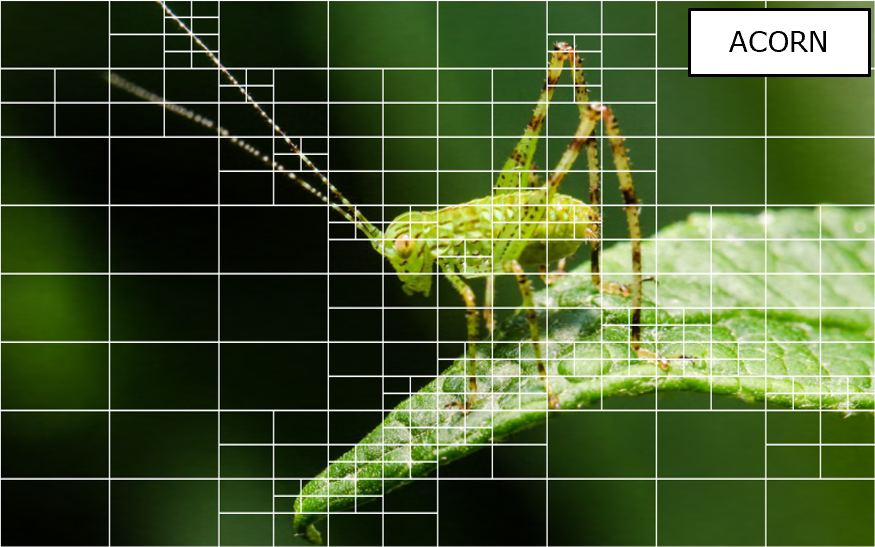}
    \caption{\textbf{MINER adaptively selects window sizes.} MINER adaptively selects the appropriate scale for each local area resulting in patch sizes that are chosen according to texture variations within the window. The figure above shows a macro photograph of a grasshopper fit by MINER (left image). Large parts of the image such as background have very smooth texture implying that they can be fit accurately at a coarser scale -- which translates to large spatial size for low frequency areas. In contrast, area around the antennae are made of high spatial frequencies, which required fitting at finer scales. ACORN provides a similar decomposition (right image) but represents image at only a single image, thereby not being amenable to multiscale analysis.}
    % \jasper{Do we need to show the ACORN image? It doesn't seem to say much}}
    \label{fig:imfit}
\end{figure}

%\begin{figure}[!tt]
%    \centering
%    \includegraphics[width=\columnwidth]{figures/imfit_timing.pdf}
%    \caption{\textbf{MINER converges $10\times$ faster than state-of-the-art approaches.} The plot shows PSNR as a function of time for various approaches for the image shown in Fig. \ref{fig:imfit}. MINER achieves higher accuracy at all times, and converges significantly faster than competing approaches. Moreover, the drop in accuracy when changing from coarse to fine scale is less severe for MINER compared to when ACORN re-estimates coordinate decomposition.}
%    \label{fig:imfit_timing}
%\end{figure}

\begin{figure}[!tt]
    \centering
    \includegraphics[width=0.45\columnwidth]{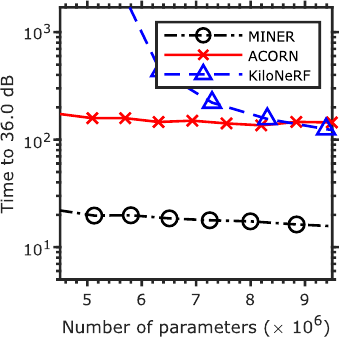}
    \hspace{2em}
    \includegraphics[width=0.435\columnwidth]{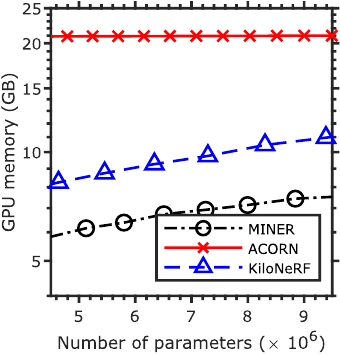}
    \caption{\textbf{MINER requires shorter training time and memory footprint.} The plot shows the time taken to achieve 36 dB and the GPU memory utilization to fit a 16MP image (Pluto) with ACORN and MINER for varying number of parameters. MINER is an order of magnitude faster than ACORN and requires less than one third of the GPU memory as ACORN -- implying MINER is well-suited to train very large models.}
    \label{fig:nparams_vs_mse}
\end{figure}
\begin{figure*}[!tt]
    \centering
    \includegraphics[width=\textwidth]{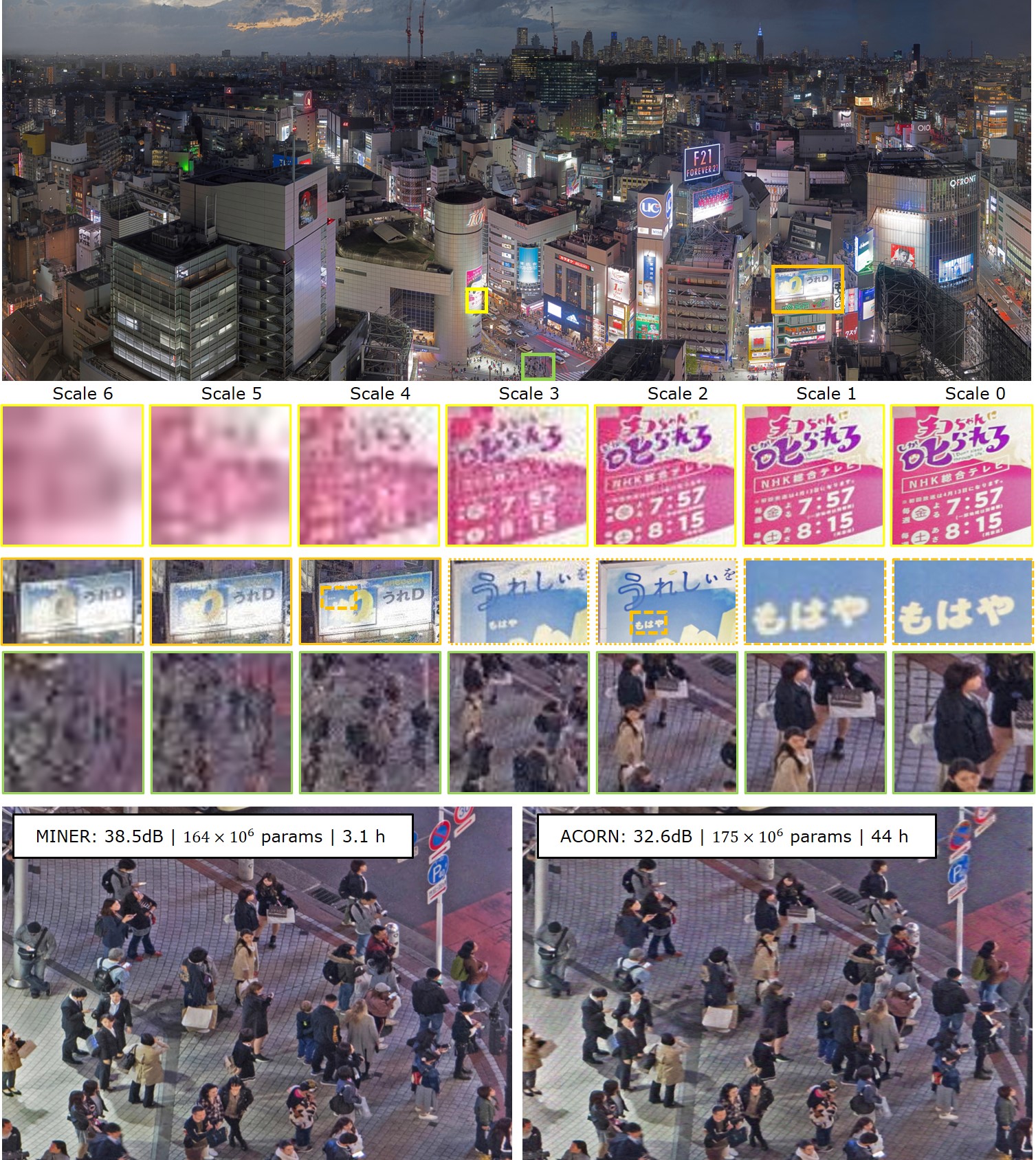}
    \caption{\textbf{Fitting gigapixel images.} The figure shows the results on fitting a gigapixel image ($20,480\times56,420$) with MINER and ACORN. MINER required 188 million parameters and converged to 38.5dB in 3.1 hours. In contrast, even after 44 hours of training, ACORN, which required 175 million parameters, achieved only 32.6dB.}
    \label{fig:gigapixel}
\end{figure*}
Figure \ref{fig:imfit_timing} shows a plot of PSNR as a function of time for various approaches.
%
%Notice the discontinuities in the training curve for MINER -- this is a due to switching from a coarser scale to finer scale, which causes a minor dip in accuracy before increasing. 
%
We also note that ACORN curve shows significant drop in accuracy as a result of re-computation of coordinate blocks. In contrast, the drop in accuracy for the MINER curve due to scale change is significantly smaller than ACORN. 
Figure \ref{fig:imfit} shows results on training a 2 megapixel image with active blocks at all scales.
The blocks are concentrated around the high frequency areas (such as the antennae of the grasshopper) as the scale increases from coarse to fine.
MINER took less than 10s to converge to 40dB fitting accuracy.
In contrast, KiloNeRF took 6  minutes to converge and ACORN took 7 minutes to converge to 40dB with approximately equal number of parameters.

Figure~\ref{fig:nparams_vs_mse} compares MINER, KiloNeRF, and ACORN in terms of time taken to achieve 36dB and GPU memory for fitting a 16MP image of Pluto.
For all three cases, we varied the number of parameters by varying the number of hidden features while keeping the number of layers fixed.
MINER consistently achieves 36dB faster than competing methods and requires significantly smaller memory footprint, making it highly scalable for large-sized problems.
\begin{figure*}[!tt]
	\centering
	\includegraphics[width=\textwidth]{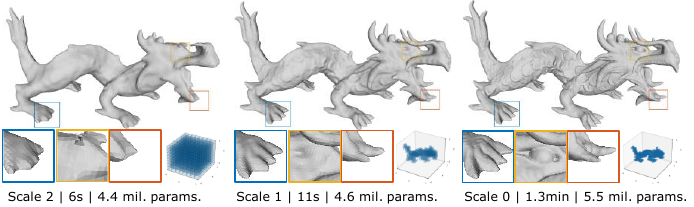}
	\caption{\textbf{Active blocks reduce with increasing scale.} The figure shows MINER results at the end of training at each scale (left column) and the active blocks at each scale (right column). As the iterations progress, only the blocks on the surface of the object remain, which leads to a dramatic reduction in non-zero blocks, and hence the total number of parameters.}
	\label{fig:active_blocks}
\end{figure*}
%
%
\begin{comment}
\begin{table}[!tt]
\caption{IOU comparison for fixed training times.}
\label{tab:IOU}
\centering
\begin{tabular}{lccc}
\toprule
& \hspace{0.2cm} \textbf{Conv. Occ.} \cite{peng2020convolutional} \hspace{0.2cm} & \hspace{0.2cm} \textbf{ACORN} \cite{martel2021acorn} \hspace{0.2cm} & \hspace{0.2cm} \textbf{MINER} \hspace{0.2cm} \\ \midrule
Conch & 0.8767 & 0.9929 & \\
Engine & 0.7043 & 0.9718 & \\
Napoleon & 0.8364 & 0.9880 & \\
Thai Statue \hspace{0.2cm} & 0.8202 & 0.9833 & \\
\bottomrule
\end{tabular}
\end{table}
\end{comment}
%
%
%\begin{table}[!tt]
%\caption{\small{\textbf{Comparison on the Thai Statue 3D Point Cloud}. MINER requires lower training time, similar GPU memory, fewer parameters, and smaller size on disk compared to state-of-the-art techniques. MINER also occupies smaller disk space compared to storing the mesh as a ply file, thereby enabling compression.}}\label{tab:comparison}
%\centering
%\begin{tabular}{lcccc}
%\toprule
          %& \hspace{0.1cm} \textbf{IOU after 10mins.} %\hspace{0.1cm} & \hspace{0.1cm} \textbf{GPU Mem.} %\hspace{0.1cm} & \hspace{0.1cm} \textbf{\# Params.} & %\hspace{0.1cm} \textbf{Size on disk} \\ \midrule
%MINER  & 0.999 & 8.5GB & 12 million & 48MB \\
%ACORN~\cite{martel2021acorn} & 0.9833 & 4.6GB & 17 million  & 68MB\\
%Conv. Occ~\cite{peng2020convolutional} & 0.8202 & 5.8GB & 0.16 million & 64KB\\
%\verb|ply| file & $-$ & $-$& $-$& 180MB
%\end{tabular}
%\end{table}
%

\begin{figure*}[!tth]
    \centering
    \includegraphics[width=0.99\columnwidth]{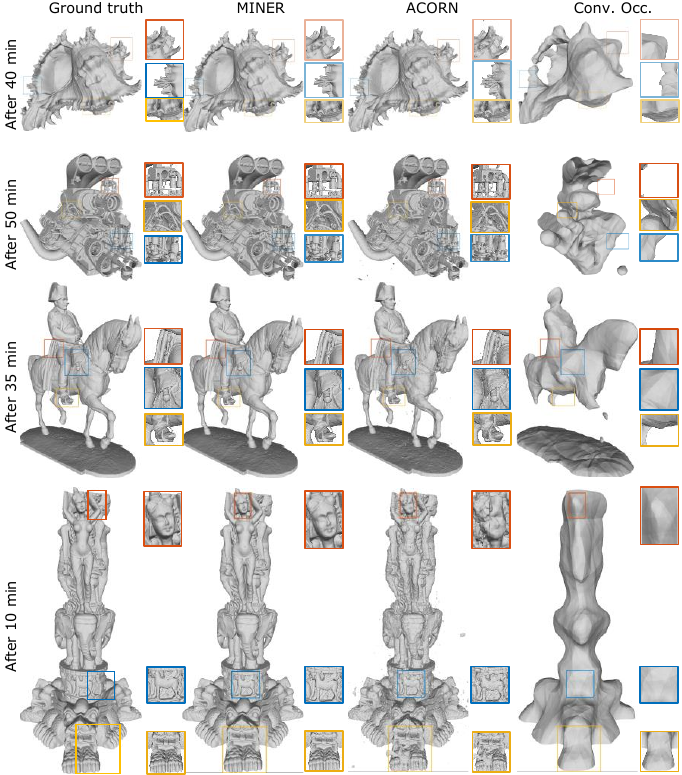}
    \caption{\textbf{Comparisons against state-of-the-art for 3D volume fitting.} We 3D occupancy fitting for a \textbf{fixed duration} with MINER, ACORN, and Convolutional occupancy~\cite{peng2020convolutional}. The number of parameters of MINER was chosen adaptively according to model complexity. MINER achieves high accuracy in a short duration for arbitrarily complex shapes, which is not possible with prior works, even though some models such as the engine (second row) require significantly more parameters.}
    \label{fig:volfit_master}
\end{figure*}

MINER scales up graciously for extremely large signals. %Figure~\ref{fig:imcompare} shows fitting result for 64 megapixel Pluto image across training iterations. 
%
%The times correspond to the instances when MINER converged at a given scale. 
%
%MINER maintains high quality reconstruction at all instances due to the multiscale training scheme and rapidly converges to a PSNR of 40dB within 50 seconds.
%
%In contrast, ACORN achieves qualitatively good results after 10s and achieves a PSNR of $30.8$ dB after 50s, and KiloNeRF achieves a qualitatively good result only after 50s.
%
%SIREN Results are not shown in the plot as the first result was produced after 4 minutes.
%
We trained ACORN on a gigapixel image shown in Fig.~\ref{fig:gigapixel} over 7 scales. We set the number of features per each block to be 9, and used a patch size of $32\times32$.
MINER converged to a PSNR of 38.5dB in 3.1h and required a total of 164 million parameters. In contrast, after 44h of training, ACORN converged to only 32.6dB while using a total of 175 million parameters.
We trained both ACORN and MINER on an 11GB NVIDIA RTX 2080 Ti GPU, which required us to decrease the maximum number of patches for ACORN to 3072 from the original authors' implementation they ran on a 48GB GPU.
MINER also enables compression of the image. Storing the image image as 16-bit \verb|tiff| format required 2.4GB of disk space. In contrast, MINER required 650 MB with 32 bit precision, implying MINER enables very high compression for images with high dynamic range.
%
%Note that although MINER requires more parameters than ACORN, the training time is shorter. Due to each MLP targeting a spatially disjoint region, the backpropagation step requires significantly lower memory, which enables MINER to train more blocks at the same time.
%
%\paragraph{Fitting CT volumes.}

\paragraph{Fitting 3D point clouds.}
%\begin{figure*}[!tt]
%    \centering
%    \includegraphics[width=\textwidth]{figures/volfit.pdf}
%    \caption{\textbf{Fitting 3D meshes.} The figure shows fitting of the Lucy 3D mesh. The top row shows active blocks at each scale, while the bottom row shows mesh obtained by running marching cube on the occupancy volume. MINER achieves a very high IoU in under 25 minutes, while ACORN takes more than 7 hours.}
%    \label{fig:volfit}
%\end{figure*}

%Implicit neural representations are best suited for high dimensional signals such as 3D meshes. 
%
Inspired by Convolutional occupancy networks~\cite{peng2020convolutional}, we utilized signed density function where the value was 1 inside the mesh and 0 outside. We sampled a total of one billion points, resulting in a $1024\times1024\times1024$ occupancy volume.
We then optimized MINER over four scales for a maximum of 2000 iterations at each scale. 
We experimented with logistic loss and MSE and found the MSE resulting in signficantly faster convergence. 
We divided the volume into disjoint blocks of size $16\times16\times16$.
The learning rate was set to $10^{-3}$. We set the number of features to 16 and the number of hidden layers to 2 for MLP for each block at all scales.
As with images, we set the per-block MSE stopping threshold to be $10^{-4}$, and did not include positional encoding for the inputs.
We then constructed meshes from the resultant occupancy volumes using marching cubes~\cite{lorensen1987marching}.
We compared our results against ACORN and convolutional occupancy networks for accuracy and timing comparisons.
For ACORN, we used the implementation and the hyperparameters provided by the original authors.
For convolutional occupancy networks, we used 200,000 randomly sampled points from the volume as input.
Comparisons against screened Poisson surface reconstruction (SPSR)~\cite{kazhdan2013screened}, which does not utilize neural networks but requires local normals, is included in the supplementary.
%For SPSR, which takes in point clouds, we sampled 5 million points from the object's surface as input.
%Note that SPSR requires extra information in the form of normals at each facet.

\begin{table}[!tt]
\caption{\small{\textbf{Comparison on the Thai Statue 3D Point Cloud}. For all experiments, we fixed the batch size to the equivalent of 1024 $16\times16\times16$ blocks. MINER requires lower training time, similar GPU memory, fewer parameters, and smaller size on disk compared to state-of-the-art techniques. MINER also occupies smaller disk space compared to storing the mesh as a ply file, thereby enabling compression.}}\label{tab:comparison}
\centering
\begin{tabular}{lcccc}
\toprule
          & \hspace{0.1cm} \textbf{IOU} \hspace{0.1cm} & \hspace{0.1cm} \textbf{GPU Mem.} \hspace{0.1cm} & \hspace{0.1cm} \textbf{\# Params.} & \hspace{0.1cm} \textbf{Size on disk} \\ \midrule
MINER - scale 3 & 0.95 (17s) & 1.8GB & 900k & 3.5MB \\
\hspace{4em} scale 2 & 0.97 (42s) & 2.5GB & 1.3 million & 4.8MB \\
\hspace{4em} scale 1 & 0.98 (1.9min) & 5.0GB & 2.8 million& 10.6MB \\
\hspace{4em} scale 0\vspace{0.5em} & 0.99 (6min) & 6.6GB & 9.9 million & 37.9MB \\
ACORN~\cite{martel2021acorn} & 0.99 (53min) & 8.0GB & 17 million & 68MB\\
Conv. Occ~\cite{peng2020convolutional} & 0.8202 & 5.8GB & 160k & 64KB\\
\verb|ply| file & $-$ & $-$& $-$& 180MB
\end{tabular}
\end{table}

Figure \ref{fig:teaser} shows the active blocks and reconstructed mesh at each scale. MINER converges in less than 25 minutes to an Intersection over Union (IoU) of 0.999. In the same time, ACORN achieved an IoU of 0.97 with worse results than MINER. ACORN took greater than 7 hours to converge to an IoU of 0.999, clearly demonstrating the advantages of MINER for 3D volumes.
We also note that MINER required less than a third as the number of parameters as ACORN -- this is a direct consequence of using block-based representation -- most blocks outside the mesh and inside the mesh converge rapidly within the first few scales, requiring far fewer representations than a single scale representation. An example of active blocks at each scale is shown in Fig.~\ref{fig:active_blocks}. As iterations progress, the number of active blocks after pruning reduce, which in turn results in more compact representation, and fewer parameters.
%
%
%Figure \ref{fig:nparams_vs_scale} shows the plot of number of cumulative parameters for each scale comparing with and without pruning. Evidently, the pruning in MINER remarkably reduces the total number of parameters without any loss in accuracy. 
%
Figure \ref{fig:volfit_master} visualizes the meshes fit with various reconstruction approaches for a fixed duration. 
The time for each experiment was chosen to be when MINER achieved an IoU of 0.999.
MINER has superior reconstruction quality compared to ACORN and convolutional occupancy networks~\cite{peng2020convolutional}. 
%
%We do note that the quality of reconstruction specifically for the engine model is superior to SPSR. Since the engine model has a large number of sharp edges, SPSR tends to oversmooth the result.
%
%Since MINER combined with marching cubes relied only on local information for reconstruction, the resultant mesh was more accurate.
%
Table~\ref{tab:comparison} compares IoU after a fixed time, GPU memory for training, number of parameters, and disk space for MINER at various scales, and competing approaches.
The memory usage of ACORN increased from 3.9GB at the start (with no further splitting) and then increased to 8GB, which we reported.
MINER achieves high accuracy (IoU) within 17s at  the coarsest scale where GPU utilization, number of parameters (and hence size on disk) are low. 
At the finest scale, MINER achieves very high accuracy, and requires fewer parameters, thereby enabling training on very large meshes.
Moreover, MINER occupies a third of the size on disk compared to a standard \verb|ply| file, thereby enabling mesh compression.

\section{Conclusions}\label{section:conclusion}
%\begin{figure}[!tt]
%    \centering
%    \includegraphics[width=\columnwidth]{figures/monalisa.pdf}
%    \caption{\textbf{MINER on very highly textured images.} Miner does not have significant benefits when the image or 3D volume has a lot of high frequency content, such as a high resolution image of the Monalisa above. We note here that the number of blocks does not reduce over scales. However, there is still a $4\times$ speed up in time compared to a single scale training (kiloNeRF) solution.}
%    \label{fig:monalisa}
%\end{figure}
%
We have proposed a novel multi-scale neural representation that trains faster, requires same or fewer parameters, and has lower memory footprint than state-of-the-art approaches.
%
%We demonstrated that the advantages of a Laplacian pyramid naturally extend to MINER.
We demonstrated that the advantages of a Laplacian pyramid including multiscale and sparse representation enable computational efficiency.
We showed that leveraging self-similarity across scales is beneficial in reducing training time drastically while not affecting the training accuracy.
MINER naturally lends itself to rendering where level-of-detail is of importance including representation and mipmapping for texture mapping.
MINER can be combined with fast, multiscale rendering approaches~\cite{yu2021plenoctrees} to achieve real time neural graphics.
With the low computational complexity and fast training and inference time, MINER opens avenues for rendering extremely large and complex geometric shapes that was previously impractical.
%
%Future directions will focus on combining MINER with multi-view synthesis to achieve extremely fast fitting and visualization pipelines that will open avenues for training highly complex geometric shapes.

%\paragraph{Limitations.} MINER has limited functionality when the texture content in the scene is extremely high.
%
%As an example, we fit MINER to a high resolution image of Monalisa in Fig.~\ref{fig:monalisa} over three scales.
%
%We note that there is no pruning of blocks at any scale, negating any gains in number of parameters.
%
%However, we still obtain gains in speed compared to a single scale training procedure such as kiloNeRF.

\section{Acknowledgements}\label{section:ack}
This work was supported by NSF grants CCF-1911094, EEC-1648451, IIS-1838177, IIS-1652633, and IIS-1730574; ONR grants N00014-18-12571, N00014-20-1-2534, and MURI N00014-20-1-2787; AFOSR grant FA9550-22-1-0060; and a Vannevar Bush Faculty Fellowship, ONR grant N00014-18-1-2047.

\balance
\bibliographystyle{splncs04}
\bibliography{refs}

\end{document}